\documentclass{article}
\usepackage{spconf,amsmath,epsfig}
\usepackage{amsmath,graphicx}
\usepackage{subfig}
\usepackage{color}
\usepackage{epstopdf}
\usepackage{tikz}
\usepackage{balance}
\usepackage{url}
\usepackage{multirow,hhline}
\usepackage{placeins}
\usepackage{wasysym}
\usepackage{amssymb}

\usepackage[nomessages]{fp}
\usepackage{array}

\newcommand{\bx}{\mathbf{x}}
\newcommand{\by}{\mathbf{y}}
\newcommand{\bw}{\mathbf{w}}
\newcommand{\bz}{\mathbf{z}}
\newcommand{\bbb}{\mathbf{b}}

\newcommand{\labsty}[1]{{\footnotesize \textbf{#1}}}
\newcommand{\Ref}{\labsty{Reference}}
\newcommand{\Noise}{\labsty{Noisy}}
\newcommand{\PPB}{\labsty{PPB}}
\newcommand{\FANS}{\labsty{FANS}}
\newcommand{\Prop}{\labsty{KL-DNN}}

\newcommand{\image}{\pgfuseimage}

\newcommand{\figpath}{./Figures/}

\definecolor{colSico}{rgb}{0.9, 0.5, 0.1}

\definecolor{mybegie}{RGB}{128,0,0}

\newcommand{\bstc}[1]{{\bf \textcolor{blue}{#1}}}

\pgfdeclareimage[width=0.13\textwidth]{ref}{\figpath 2000.jpg}
\pgfdeclareimage[width=0.13\textwidth]{speckle}{\figpath f.png}
\pgfdeclareimage[width=0.13\textwidth]{noisy}{\figpath 02000_1_speckled.png}

\pgfdeclareimage[width=0.18\columnwidth]{ref1}{\figpath 1408_ref.png}
\pgfdeclareimage[width=0.18\columnwidth]{noisy1}{\figpath 1408_noisy.png}
\pgfdeclareimage[width=0.18\columnwidth]{ppb1}{\figpath 1408_ppb.png}
\pgfdeclareimage[width=0.18\columnwidth]{fans1}{\figpath 1408_fans.png}
\pgfdeclareimage[width=0.18\columnwidth]{L21}{\figpath 1408_L2.png}
\pgfdeclareimage[width=0.18\columnwidth]{L2KL1}{\figpath 1408_L2_KL.png}

\pgfdeclareimage[width=0.18\columnwidth]{ref2}{\figpath 1405_ref.png}
\pgfdeclareimage[width=0.18\columnwidth]{noisy2}{\figpath 1405_noisy.png}
\pgfdeclareimage[width=0.18\columnwidth]{ppb2}{\figpath 1405_ppb.png}
\pgfdeclareimage[width=0.18\columnwidth]{fans2}{\figpath 1405_fans.png}
\pgfdeclareimage[width=0.18\columnwidth]{L2KL2}{\figpath 1405_L2_KL.png}

\pgfdeclareimage[width=0.15\textwidth]{real}{\figpath vele_noisy.png}
\pgfdeclareimage[width=0.15\textwidth]{real_ppb}{\figpath vele_ppb.png}
\pgfdeclareimage[width=0.15\textwidth]{real_fans}{\figpath vele_fans.png}
\pgfdeclareimage[width=0.15\textwidth]{real_prop}{\figpath vele_L2_KL.png}

\pgfdeclareimage[width=0.25\columnwidth]{real2}{\figpath vele_noisy_2.png}
\pgfdeclareimage[width=0.25\columnwidth]{real_ppb2}{\figpath vele_ppb3.png}
\pgfdeclareimage[width=0.25\columnwidth]{real_fans2}{\figpath vele_fans_2.png}
\pgfdeclareimage[width=0.25\columnwidth]{real_prop2}{\figpath vele_L2_KL_2.png}

\pgfdeclareimage[width=0.15\textwidth]{real1}{\figpath vele_noisy_2.png}
\pgfdeclareimage[width=0.15\textwidth]{ratio_ppb}{\figpath ratio_vele_ppb.png}
\pgfdeclareimage[width=0.15\textwidth]{ratio_fans}{\figpath ratio_vele_fans.png}
\pgfdeclareimage[width=0.15\textwidth]{ratio_prop}{\figpath ratio_vele_L2_KL.png}

\pgfdeclareimage[width=0.25\columnwidth]{ratio_ppb2}{\figpath ratio_vele_ppb.png}
\pgfdeclareimage[width=0.25\columnwidth]{ratio_fans2}{\figpath ratio_vele_fans.png}
\pgfdeclareimage[width=0.25\columnwidth]{ratio_prop2}{\figpath ratio_vele_L2_KL.png}

\pgfdeclareimage[width=0.9\columnwidth]{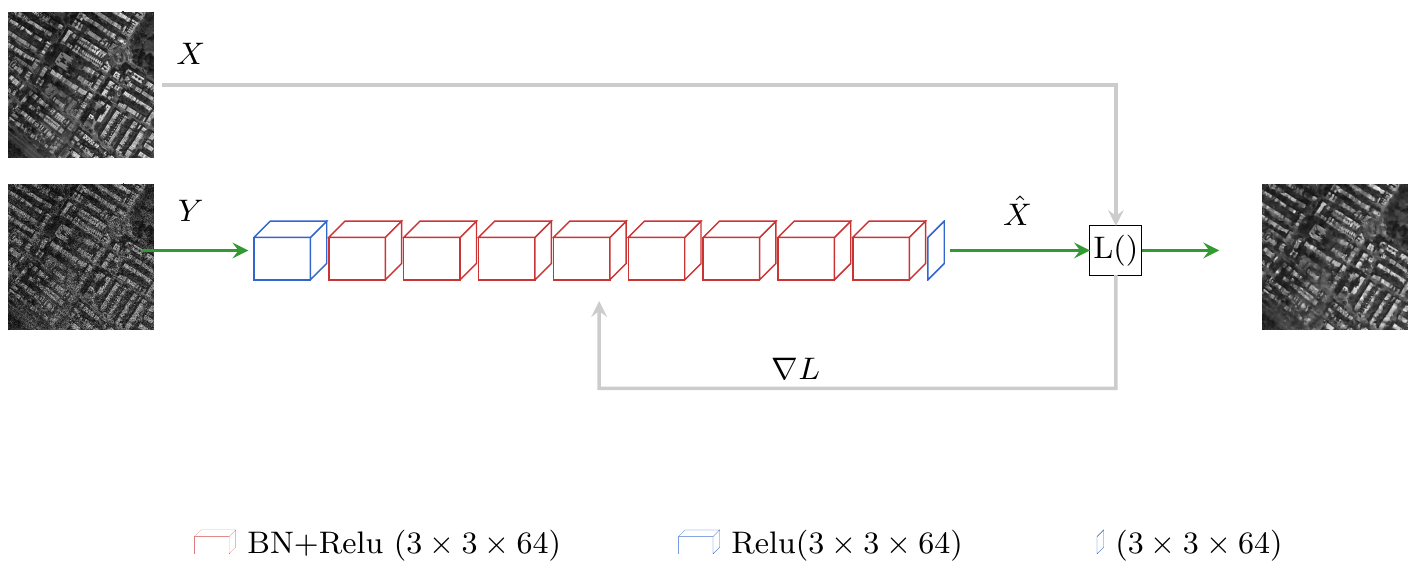}{\figpath net_1.pdf}

\title{A New Ratio Image Based CNN algorithm for SAR Despeckling}
\name{Sergio Vitale $^{1}$, Giampaolo Ferraioli $^{2}$ and Vito Pascazio $^{1}$}
\address{$^{1}$ Dipartimento di Ingengeria, Universit\`{a} di Napoli Parthenope \\ $^{2}$Dipartimento di Scienze e Tecnologie, Universit\`{a} di Napoli Parthenope}
\begin{document}
	%
	\maketitle

	\begin{abstract}
		In SAR domain many application like classification, detection and segmentation are impaired by speckle. Hence, despeckling of SAR images is the key for scene understanding. Usually despeckling filters face the trade-off of speckle suppression and information preservation. In the last years deep learning solutions for speckle reduction have been proposed. One the biggest issue for these methods is how to train a network given the lack of a reference. In this work we proposed a convolutional neural network based solution trained on simulated data. We propose the use of a cost function taking into account both spatial and statistical properties. The aim is two fold: overcome the trade-off between speckle suppression and details suppression; find a suitable cost function for despeckling in unsupervised learning. The algorithm is validated on both real and simulated data, showing interesting performances.
	\end{abstract}
	\section{Introduction}
	\label{sec:intro}
	Interpretation and understanding of remote sensing images is always an open issue. There are a lot of applications such as object detection, classification, land use segmentation and denoising aiming to extract information by remote sensing data. A very challenging environment is the Synthetic Aperture Radar (SAR) imaging system. SAR images are affected by multiplicative noise, called speckle, that impairs the performance in all applications. In the last decades several despeckling filters have been proposed. Generally, even if there is not a clear-cut classification, the speckle filters are divided in two groups: local and non local filters.
	The formers produce filtered images where each output pixel is given by averaging values in its neighbourhood, assuming that closer pixels should bring similar information. Filters like Lee, its enhanced version and Kuan filter belong to this class. These filters suffer of edges blurring. On the edge between two structures the pixel values can be very different and averaging the neighbourhood produces smoothness in the filtered image. In order to overcome this issue, non local filters like PPB, SAR-BM3D, FANS and NL-SAR look for similarity in a larger windows search instead of a close neighbourhood. For a review of former method refer to \cite{Argenti2013}, instead for latters to \cite{Deledalle2014} and for FANS to \cite{Cozzolino2014}.
	In the last years, deep learning is spreading in several image processing fields achieving very good results, not less in the remote sensing community. Deep learning methods have been proposed in applications like classification, detection, data fusion and despeckling. In \cite{Chen2014} a CNN based approach for SAR target detection is proposed. In \cite{Scarpa2018} and \cite{Scarpa2018RS} deep learning methods for pansharpening and SAR-optical data fusion are used. 
	
	Regarding despeckling, the lack of a clean reference is still an open issue. In order to overcome this problem, \cite{Wang2017} train a CNN using simulated data, instead Chierchia et al \cite{Chierchia2017} trains the network on multilook real SAR images.
	
	Following \cite{Wang2017}, in this work we propose a supervised CNN for despeckling, using a cost function given by a linear combination of a per-pixel and a statistical loss. The aim is to show how extracting statistical information helps the network to solve the usual trade-off between speckle suppression and details preservation. Moreover, the use of a statistical loss can help in future to build a neural network for unsupervised despeckling.

	\section{Convolutional Neural Networks}
	
	As said, deep learning and convolutional neural networks are becoming fundamental part of several image processing applications.
	There is not a predefined structure for a CNN, but in general it is a neural network that, in addition or substitution of fully connected layers use convolutional layers. 
	Generally, different layers are combined: convolutional, pooling, batch-normalization, soft-max, non linearities.
	The number, the kind and the way in which they are combined depend on the final task.
	
	A generic convolutional layer is define by a set of $ M $ kernels of dimension $(K \times K)$. 
	So the $l$-th generic convolutional layer,  for $N$-bands input $\bx^{(l)}$, yields an $M$-band 
	output $\bz^{(l)}$ 

	\[
	\bz^{(l)} = \bw^{(l)} \ast \bx^{(l)} + \bbb^{(l)},
	\]
	\\
	whose $m$-th component is a combination of 2D convolutions:
	
	\[
	\bz^{(l)}(m,\cdot,\cdot) = \sum_{n=1}^N  \bw^{(l)}(m,n,\cdot,\cdot) \ast \by^{(l)}(n,\cdot,\cdot)+ \bbb^{(l)}(m).
	\]

	The tensor $\bw$ is a set of $M$ convolutional $(K\times K)$ kernels, while $\bbb$ is a $M$-vector bias.
	
	Let $\Phi_l\triangleq\left(\bw^{(l)},\bbb^{(l)}\right)$ be the learnable parameters of $l$-th layer.
	Usually, the output of the layer is followed by an activation function $g_l(\cdot)$ in order to introduce non-linearities.
	In this work all the convolutinal layers, except the first and the last, are followed by a pointwise ReLU activation function $g_l(\cdot)\triangleq \max(0,\cdot)$ producing the intermediate layer outputs ( the set  of  $ M $ so-called {\em feature maps})

	\[
	\by^{(l)}
	\triangleq f_l(\bx^{(l)},\Phi_l) =
	\begin{cases}
	\max(0,\bw^{(l)} \ast \bx^{(l)} + \bbb^{(l)}), & l<L\\
	\bw^{(l)} \ast \bx^{(l)} + \bbb^{(l)}, &  l=L
	\end{cases}
	\]
	whose concatenation gives the overall CNN function
	\\
	\begin{equation}
	f(\bx,\Phi) = f_L(f_{L-1}(\ldots f_1(\bx,\Phi_1),\ldots,\Phi_{L-1}),\Phi_L)
	\nonumber
	\label{eq:chain}
	\end{equation}
	\\
	where $\Phi\triangleq(\Phi_1,\ldots,\Phi_L)$ is the whole set of parameters to learn.
	
	Inorder to train the parameters, several training couples of input and reference output samples must be provided, a cost function and a optimization process must be chosen.
	The cost function $L(\cdot,\Phi)$ compares the similarity between predicted and reference outputs. An optimization process tries to minimize $L$ and depending on it the parameters $\Phi$ are updated.
	
	\section{PROPOSED METHOD}

	\label{sec:print}
	\subsection{Speckle Data Simulation}
	As said before, SAR images are affected by a multiplicative noise called speckle.  
	Let $Y$ be a intensity SAR image, it can be expressed as \cite{Argenti2013}:
	\\
	\begin{equation}
	Y = f(X,N) = X\cdot N
	\label{eq: sar formation}
	\end{equation}
	\\
	where $X$ is the noise-free image and $N$ is the multiplicative speckle. In the hypothesis of fully developed speckle, $N$ has a Gamma distribution \cite{touzi2002}:
	\\
	\begin{equation}
	p(N) = \frac{1}{\Gamma(L)} N^L e^{-NL}
	\label{eq: speckle distribution}
	\end{equation}
	\\
	where $L$ is the number of looks of the SAR image.
	An ideal despeckling filter will remove the noise without introducing artefacts and preserving the spatial informations.
	
	In this work simulated data are used following the scheme in equations (\ref{eq: sar formation})-(\ref{eq: speckle distribution}). We consider three datasets of clean images \cite{Wang2017}: scraped Google Maps that provides urban images, UCID and BSD  that provide generic images. We simulated speckle with Gamma distribution and apply it on this three datasets.
	
	\subsection{Training with Kullback-Leibler divergence}
	 	\begin{figure}[h]
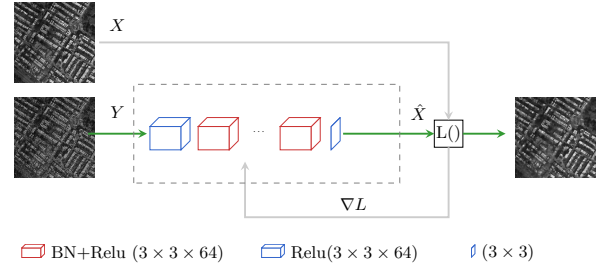

	 		\centering
	 		\image{net}
	 		\caption{CNN architecture}
	 		\label{fig: net}
	 	\end{figure}
	 	
	In the proposed method the network (Fig.\ref{fig: net}) is composed by 10 convolutional layers. In order to speed up the convergence, each layer except the first and last is followed by a Rectified Linear Unit (ReLu). Moreover for the same reason, batch normalization is performed for each layer except the last.
	The training process is performed by the Stochastic Gradient Descent with momentum, with learning rate $ \eta = 2 \cdot 10^{-6}$ on $30000 \times (65 \times 65)$ training patches and $12000 \times (65 \times 65)$ for the validation.
	
	In this work we focus our attention on a customized cost function which aim is two fold: firstly have better speckle suppresion and edge preservation; secondly moving towards unsupervised despeckling.
	
	Given a single band noisy image $Y$, and the noise-free reference image $X$,
	the predicted output is  $ \hat X=f(\bx,\Phi)$ (see in Fig.\ref{fig: net}) and  the predicted noise is $\hat{N} = Y/\hat{X}$ 

	In this work we propose as cost function $L$ a linear combination given by 
	
	\begin{equation}
	L(x,\Phi) = ||f(x,\Phi) - X||_2^2 + \lambda \sum_{i} \log_2 \frac{p_{\hat{N}}(i)}{p_N(i)} \cdot p_{\hat{N}}(i)
	\label{eq: cost}
	\end{equation}
	
	where the first term is the Mean Square Error (MSE) between output and its reference, while the second one is the Kullback-Leibler divergence (KL) between the predicted noise probabilities distribution $p_{\hat{N}}$  and that of simulated speckle noise $p_N$ (that follow the Gamma distribution)
	
	
	
	With this cost function the network predicts the clean image taking into account the statistical speckle properties. MSE forces the network to predict directly the image by a per pixel comparison with the reference, and KL ensures that the removed noise has probability distributions as close as possible to the Gamma distribution.
	
	As said in previous sections, all the methods show a trade-off between speckle reduction and edge preservation. With the combination in equation (\ref{eq: cost}) we try to preserve both spatial and spectral informations.
	Moreover, using a cost function like $L_2$ means to use a cost function that is independent from the reference and it can be an elegible cost function for despeckling in unsupervised neural networks. Based on the adopted methodologies, the proposed technique will be referred as Kullbacl-Leibler Despeckling Neural Network (KL-DNN) algorithm
		\begin{figure}[h]
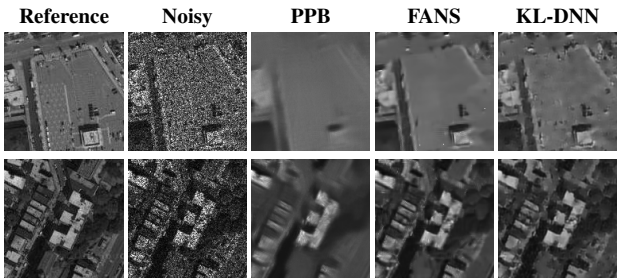

			\centering
			\begin{tabular}{ccccc}
				\Ref & \Noise & \PPB & \FANS & \Prop\\
				\image{ref1} & \image{noisy1} & \image{ppb1} &	\image{fans1}  & \image{L2KL1}\\
				\image{ref2} & \image{noisy2} & \image{ppb2} &	\image{fans2}  & \image{L2KL2}\\
			\end{tabular}
			\caption{Results on simulated images: clip1 (above), clip2 (bottom)}
			\label{fig: res_simul}
		\end{figure}

	\section{Experiments}
		\begin{figure*}[h]
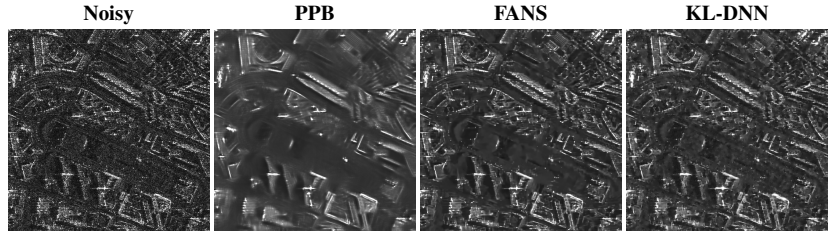

			\centering
			\begin{tabular}{cccc}
				\Noise & \PPB & \FANS & \Prop\\
				\image{real} & \image{real_ppb} &	\image{real_fans}  & \image{real_prop}\\
			\end{tabular}
			\caption{Results on real images: the first image on the left is the SAR real image.  }
			\label{fig: res_vele}
		\end{figure*}
		
		\begin{figure*}[h]
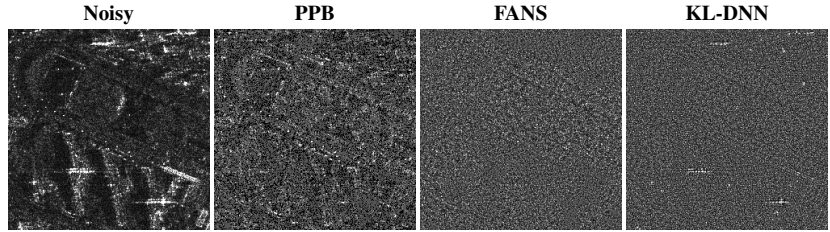

			\centering
			\begin{tabular}{cccc}
				\Noise & \PPB & \FANS & \Prop\\
				\image{real1} & \image{ratio_ppb} &	\image{ratio_fans}  & \image{ratio_prop}\\
			\end{tabular}
			\caption{Ratio images referred to the central patch of the image: first image on the left is the real SAR patch }
			\label{fig: ratio_vele}
		\end{figure*}
%
%
%

	The proposed method is tested on both simulated and real SAR images and a comparison two of the most accredited despeckling algorithms, PPB \cite{Deledalle2009} and FANS \cite{Cozzolino2014}, is conducted. 
	Fig. \ref{fig: res_simul} shows the results on two simulated clips. KL-DNN seems to have better performance compared with PPB and FANS. PPB tends to be oversmoothed on both clips loosing a lot of spatial details. FANS preserves the details better respect PPB but suffers of oversmoothing mainly on small objects. For example, in the first clip FANS misses all the cars in the parking lot, instead in the second clip the edge of the buildings are blurred. Moreover some distortions appear on the roof.
	Regarding KL-DNN, on clip1 it better preserves small objects like cars. Moreover the image seems more detailed than PPB and FANS. In clip2 KL-DNN has similar performances to FANS, even if the edge of building and spatial details like foliages of trees are better preserved.
	The disadvantage of KL-DNN is that it suffers of distortion on smooth surfaces: in the parking lot and on the building's roof some dark spots appear.
	
	In order to have a numerical assessment, for the simulated data in which the clean reference is available, the SSIM, PSNR and SNR indexes are computed.
	Tab. \ref{tab: res} confirms the previous consideration: KL-DNN shows an improvement with respect to PPB and FANS mainly on the first clip; regarding the second clip the gain respect FANS is more slight.
	
%
	
	In Fig. \ref{fig: res_vele}, results on real SAR image are shown. The image presents both man-made objects and natural areas.

	Generally, the high value given by multiple reflections in man-made areas are very challenging to filter and all the three methods shows some difficulties. 
	PPB filters the big scale object like building but it tends to be over smoothed in flat zones, suppressing a lot of details. FANS and KL-DNN have very close performance: FANS seems to better preserve building details, however it introduces some artefacts and oversmooths homogeneous areas; on the other hand KL-DNN finds challenging the man made structures while showing good performances in other areas.\\
	To have a more clear idea of the filtering quality, ratio images of the a central patch of image are shown in Fig. \ref{fig: ratio_vele}. The ratio image of a perfect filter should contain only speckle. Considering the ratio image of PPB, it is evident that PPB suppresses too many details.
	As we expect, KL-DNN faces difficulties with strong backscattering (multiple bouncing). However, it is important to note that, except for these points, the ratio image of KL-DNN appears characterized by homogeneously distributed speckle, differently from FANS where some structures appear. It means that KL-DNN has a better ability in suppressing speckle while preserving the details.\\
	In order to have a numerical assessment the M-index \cite{Gomez2017} and KL divergence are listed in Tab.\ref{tab: res}. KL-DNN achieves the best performance for both indexes, validating the qualitative analysis.

	\begin{table}[h]
		\centering
		\setlength\tabcolsep{5 pt}
		\renewcommand{\arraystretch}{1.1}
		\begin{tabular}{c}
			\begin{tabular}{lccc}
				
				\hline
				\setlength\tabcolsep{1 pt}
				& SSIM 	& PSNR & SNR \\
				\hline  
				PPB  & 0.6715	& 24.9778 & 5.4899\\
				FANS &  0.7731	& 27.2326 & 7.7447\\
				KL-DNN & \bstc{0.7890}	& \bstc{27.5075} & \bstc{8.0195}\\
				\hline
			\end{tabular}
			\\
			a)
			\\
			\begin{tabular}{lccc}
				
				\hline
				\setlength\tabcolsep{1 pt}
				& SSIM 	& PSNR & SNR\\
				\hline
				PPB  & 0.6356	& 23.0970 & 6.0597\\
				FANS &  0.762	& 25.6149 & 8.5775\\
				KL-DNN & \bstc{0.7725}	& \bstc{25.7524} & \bstc{8.7151}\\
				\hline
			\end{tabular}
			\\
			b)
			\\
			\begin{tabular}{lcccc}
				\hline
				\setlength\tabcolsep{1 pt}
				\renewcommand{\arraystretch}{1.1}
				& M-index & &&KL div\\
				\hline
				PPB & 12.04&&&0.0147 \\
				FANS &  11.22&&&0.0089\\
				KL-DNN & \bstc{9.93}&&& \bstc{0.0081}\\
				\hline
			\end{tabular}
			\\
			c)\\
			
		\end{tabular}
		
		\caption{Numerical Results: a) evaluation on simulated clip1. b) evaluation on simulated clip2. c) evaluation on real SAR image}
		\label{tab: res}
	\end{table}


	\section{Conclusions}

%
%

	\label{sec:foot}
	In this work a convolutional neural network for despeckling is proposed. We use a cost function that relies on per-pixel distance between output and reference and, at the same time, on the statistical properties of the noise. The use of this cost function helps the network to suppress the noise while preserving spatial details. Despite some spatial distortions, the proposed method seems to have better detail preservation than other methods, mainly on small object. In fact, more small details are preserved which is a key feature for the scene interpretation of the remote sensing image.
	\bibliographystyle{IEEEbib}
	\bibliography{refs}
	
\end{document}